\def\year{2022}\relax
\documentclass[letterpaper]{article} 
\usepackage{aaai22}  
\usepackage{times}  
\usepackage{helvet}  
\usepackage{courier}  
\usepackage[hyphens]{url}  
\usepackage{graphicx} 
\usepackage{natbib}  
\usepackage{caption} 
\DeclareCaptionStyle{ruled}{labelfont=normalfont,labelsep=colon,strut=off} 
\usepackage{algorithm}
\usepackage{algorithmic}

\usepackage{newfloat}
\usepackage{listings}
\floatstyle{ruled}
\newfloat{listing}{tb}{lst}{}
\floatname{listing}{Listing}
\title{Causal Intervention for Subject-Deconfounded Facial Action Unit Recognition}
\author {
    Yingjie Chen\textsuperscript{\rm 1},
    Diqi Chen\textsuperscript{\rm 2},
    Tao Wang\textsuperscript{\rm 1}\thanks{Corresponding author.},
    Yizhou Wang\textsuperscript{\rm 1},
    Yun Liang\textsuperscript{\rm 1}
}
\affiliations {
    \textsuperscript{\rm 1} School of Computer Science, Peking University, Beijing China \\
    \textsuperscript{\rm 2} Advanced Institute of Information Technology (AIIT), Peking University, Hangzhou, China\\
    chenyingjie@pku.edu.cn, dqchen@aiit.org.cn, wangtao@pku.edu.cn, yizhou.wang@pku.edu.cn, ericlyun@pku.edu.cn
}

\usepackage{amssymb}
\usepackage{amsthm,amsmath}
\usepackage{mathrsfs}
\usepackage{multirow}
\usepackage{booktabs}
\usepackage{color}
\usepackage{bbm}
\usepackage{bbding}
\usepackage{threeparttable}
\usepackage{xspace}

\begin{document}

\makeatletter
\DeclareRobustCommand\onedot{\futurelet\@let@token\@onedot}
\def\@onedot{\ifx\@let@token.\else.\null\fi\xspace}

\def\eg{\emph{e.g}\onedot} \def\Eg{\emph{E.g}\onedot}
\def\ie{\emph{i.e}\onedot} \def\Ie{\emph{I.e}\onedot}
\def\cf{\emph{c.f}\onedot} \def\Cf{\emph{C.f}\onedot}
\def\st{\emph{s.t}\onedot} \def\Cf{\emph{S.t}\onedot}
\def\etc{\emph{etc}\onedot} \def\vs{\emph{vs}\onedot}
\def\wrt{w.r.t\onedot} \def\dof{d.o.f\onedot}
\def\etal{\emph{et al}\onedot}
\makeatother

\maketitle

\begin{abstract}
Subject-invariant facial action unit (AU) recognition remains challenging for the reason that the data distribution varies among subjects.
In this paper, we propose a causal inference framework for subject-invariant facial action unit recognition.
To illustrate the causal effect existing in AU recognition task, we formulate the causalities among facial images, subjects, latent AU semantic relations, and estimated AU occurrence probabilities via a structural causal model. By constructing such a causal diagram, we clarify the causal effect among variables and propose a plug-in causal intervention module, CIS, to deconfound the confounder \emph{Subject} in the causal diagram. 
Extensive experiments conducted on two commonly used AU benchmark datasets, BP4D and DISFA, show the effectiveness of our CIS, and the model with CIS inserted, CISNet, has achieved state-of-the-art performance.

\end{abstract}

\begin{figure}[!t]
    \centering
    \includegraphics[scale=0.28]{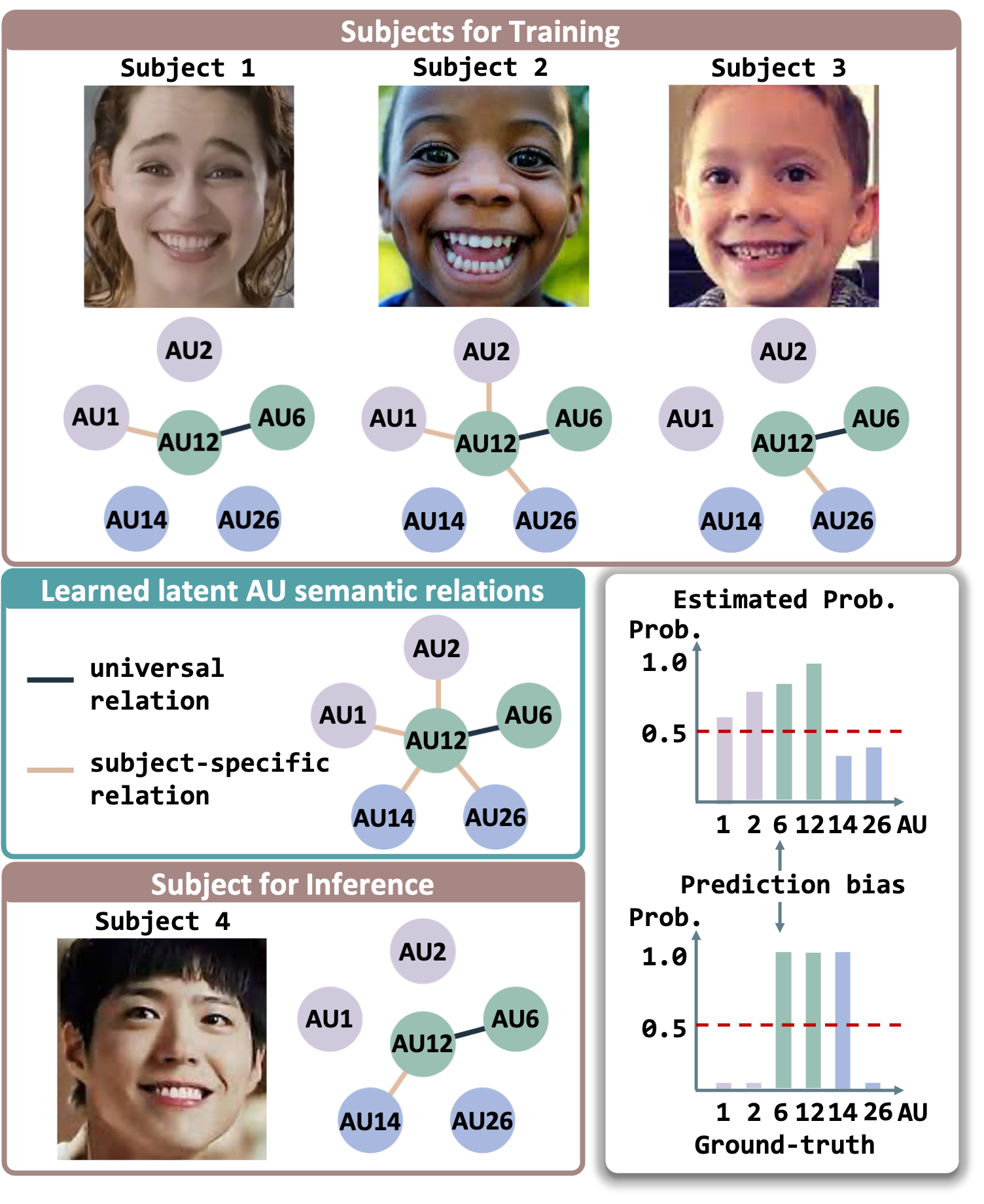}
    \caption{Illustration of subject variation problem. The AU semantic relations embedded in the facial images of the four subjects vary due to the differences among their customs of expressing happiness. Using samples collected from the first three subjects for training may cause the learned latent AU semantic relations containing subject-specific ones (dark yellow lines) in addition to universal ones (dark blue lines). When encountering a new subject (\emph{Subject 4}) in the inference stage, differences between \emph{Subject 4}'s specific AU relations and the learned ones will lead to prediction bias.}
    \label{fig:motivation}
\end{figure}

\section{Introduction}
With the proliferation of facial behavior analysis in real-world application scenarios such as online education and driver safety assistance, facial action unit recognition has attracted increasing research interest as a fundamental task in the field of affective computing. According to Facial Action Coding System (FACS)~\citep{FACS}, facial action units (AUs), defined as the combinations of facial muscle movements, can describe almost all facial behaviors, which is essential for fine-grained facial behavior analysis. 
In recent years, deep learning has proved its efficacy and efficiency in facial action unit recognition task~\citep{cui2020knowledge,chen2021cross,yang2021exploiting,song2021hybrid}, but there is still room for improvement since some inherent nature of AU has not been fully exploited.

AUs are not independent of each other.
On the one hand, AUs usually do not occur alone when humans express certain emotions, and thus some combinations of AUs, which pertain to displayed emotions, can be frequently observed, \eg~AU6 (Cheek Raiser) and AU12 (Lip Corner Puller) tend to appear together and form facial expression \emph{happiness}~\citep{basic-expression}. On the other hand, there are strict co-occurrence and mutual exclusion among AUs due to the structural constraints brought by facial anatomy, \eg~AU22 (Lip Funneler) and AU23 (Lip Tightener) cannot appear simultaneously since they are all related to facial muscle \emph{Orbicularis oris}, and it is difficult to make AU9 (Nose Wrinkler) without the presence of AU6 (Cheek Raiser) due to muscular synergy~\citep{zarins2018anatomy}.
Although the presence or absence of each AU can be mainly inferred from facial appearance changes, it can also be partially inferred based on the states of other AUs. Therefore, an accurate AU recognition model captures not only low-level facial appearance features but also high-level semantic relations among AUs.

Recent works have made progress in capturing high-level AU semantic relations in an implicit way~\citep{DSIN_2018_Corneanu,LP_2019_CVPR} by exploiting correlations between AUs via probabilistic graphic models or in an explicit way~\citep{SRERL_li2019semantic,shao2020spatio} by constructing an AU semantic graph according to statistics of the training data, and both kinds of works have achieved more accurate AU recognition. 
Although these methods can make use of priors contained in the training data, they all suffer from prediction bias while being applied to samples of new subjects.
This is known as subject variation problem, which makes it challenging for AU recognition models to generalize across subjects.
Although previous works have noticed that subject variation problem exists in facial action unit recognition task, as far as we know, there have been few works focusing on answering the whys and wherefores. 

We argue that the prediction bias caused by subject variation problem is mainly due to the fact that the latent AU semantic relations vary among subjects.
As shown in Fig.~\ref{fig:motivation}, subjects in all facial images are expressing the facial expression \emph{happiness}, which is composed of AU6 and AU12 as mentioned in~\citep{basic-expression}. Thus the co-occurrence of AU6 and AU12 is one universal AU semantic relation shared by all subjects. However, AU semantic relations embedded in each facial image vary among subjects, which means that in addition to the universal relation, there are subject-specific AU semantic relations due to the differences in subjects' customs of expressing emotions. \Eg~\emph{Subject 1} tends to raise her inner brows (AU1) while smiling and \emph{Subject 2} tends to raise his whole brows (AU1 and AU2) and laugh with the drop of his jaw (AU26).
If we train a model using samples of the first three subjects, the model will learn a set of latent AU semantic relations containing both the universal ones and the subject-specific ones. When applying the model to \emph{Subject 4}, the subject-specific relations of the training subjects may lead to prediction bias on AU1, AU2, or AU26.


So far, we can see that the prediction bias of AU recognition is mainly caused by the differences among subjects' customs of expressing emotions. \emph{Subject} can be essentially regarded as a confounder, which misleads AU recognition model to learn subject-specific AU semantic relations from subjects in the training data and thus causes prediction bias while applying the model to a new subject for inference.
For clarity, we denote the input facial images as $X$ and the predicted AU occurrence probabilities as $Y$, and an AU recognition model aims to approximate $P(Y|X)$ as much as possible. 
However, as mentioned above, $P(Y|X)$ may lead to a biased AU recognition model since it may learn subject-specific AU relations that are not shared by new subjects. For example, $P(Y|X)$ would learn the relation that AU2 and AU12 tend to co-occur when using samples of \emph{Subject 1} in Fig.~\ref{fig:motivation} for training, which is actually a subject-specific relation of \emph{Subject 1} which is not suitable for others, and $P(Y|X)$ would mistakenly reduce the importance of the relation that AU6 and AU12 co-occur and form the facial expression \emph{happiness} when using facial images of \emph{Subject 3} for training, since \emph{Subject 3} may have physical difficulty in contracting facial muscles related to the occurrence of AU6. To relieve subject variation problem, we propose a method to learn the universal AU semantic relations by making our model approximate $P(Y|do(X))$ instead of $P(Y|X)$, where $do$-operation denotes the pursuit of the causality between the cause $X$ and the effect $Y$ without the confounding effect caused by the confounder \emph{Subject}.

To this end, we formulate subject variation problem by constructing a causal diagram to analyze the causalities among facial images, subjects, latent AU semantic relations, and estimated AU occurrence probabilities. Our causal inference framework not only fundamentally explains how subject-specific AU semantic relations hurt the performance of AU recognition models, but also provides a solution by removing the effect caused by confounder \emph{Subject}. Based on the causal model, a plug-in causal intervention module called \textbf{CIS} is proposed to deconfound \emph{Subject} via back-door adjustment~\citep{pearl2016causal}. 

Our main contributions are listed as:
\begin{itemize}
    \item We formulate subject variant problem in AU recognition using an AU causal diagram to explain the whys and wherefores. To the best of our knowledge, this is the first work to explain this problem with the help of causal inference theory and make attempt to remove the effect caused by subject variation via causal intervention.
    \item Based on our causal diagram, we propose a plug-in causal intervention module, CIS, which could be inserted into advanced AU recognition models for removing the effect caused by confounder \emph{Subject}.
    \item Extensive experiments on two widely used AU benchmark datasets, BP4D and DISFA, demonstrate that the proposed CIS can boost various AU recognition models to new state-of-the-art.
\end{itemize}

\section{Related Work}
\subsection{Facial Action Recognition}
In recent years, research on facial action unit recognition has seen great achievements. The rise of deep learning has raised the performance of AU recognition to a new level. 
Considering the locality of AUs, methods such as~\citep{DRML_2016_CVPR,ROINet_2017_CVPR,EAC-Net,song2021uncertain,chen2021cafgraph} make attempt to learn better facial appearance features by emphasizing important local facial regions.
Zhao~\etal~\citep{DRML_2016_CVPR} proposed Deep Region and Multi-label Learning (DRML), which employs a region layer to induce important facial regions and force the learned weights to capture structural information of the face. 
Considering the semantic relations among AUs, some works~\citep{wang2013capturing,walecki2017copula} make efforts in modeling such relations via probabilistic graphical models or graph neural networks. Wang~\etal~\citep{wang2013capturing} introduced a restricted Boltzmann machine to model facial action units, thereby capturing not only local but also global AU dependencies.
Li~\etal~\citep{SRERL_li2019semantic} investigated how to integrate the semantic relationship propagation between AUs to enhance the feature representation of facial regions, and proposed an AU semantic relationship embedded representation learning (SRERL) framework. 
However, these works ignore subject variation problem and thus the obtained AU recognition models suffer from the subject-related prediction bias.

As for subject variation problem, works such as~\citep{chen2013learning} provide a solution for enhancing the generalizability of AU recognition model by training personalized AU classifiers for each subject and works such as~\citep{zen2016learning,GARN_2018_ACMMM} make attempt to relieve the subject-related prediction bias through domain adaptation.
Although these works have realized that the data distribution of training subjects differs from that of unseen subjects, they are still based on the assumption that the data distribution of source and target domains shares some similarities. 
In contrast, we formulate the causalities among variables in AU recognition task via a structural causal model to answer the whys and wherefores of subject variation problem and provide a solution based on causal intervention. 

\subsection{Causal Inference in Computer Vision}
Causal inference~\citep{pearl2000models,rubin2005causal} has been gradually applied to computer vision tasks in recent years, such as long-tailed classification~\citep{tang2020long}, weakly-supervised semantic segmentation~\citep{zhang2020causal}, few-shot learning~\citep{yue2020interventional}, and class-incremental learning~\citep{hu2021distilling}. Causal inference empowers models the capability to consider the causal effect that naturally exists in a task and disentangle direct effect and indirect effect. There are two ways for causal inference: one is Pearl’s structural causal model~\citep{pearl2000models}, and the other is the potential outcome framework proposed by Robins and Greenland~\citep{rubin2005causal}, in terms of which they set out to express their conception of confounding. The advantage of structural causal model is that it shows the causal and effect among several variables in the form of a causal diagram, which is more intuitive and conductive for analysis, and thus we choose to follow the first way in our work.

\begin{figure}[!t]
    \centering
    \includegraphics[scale=0.43]{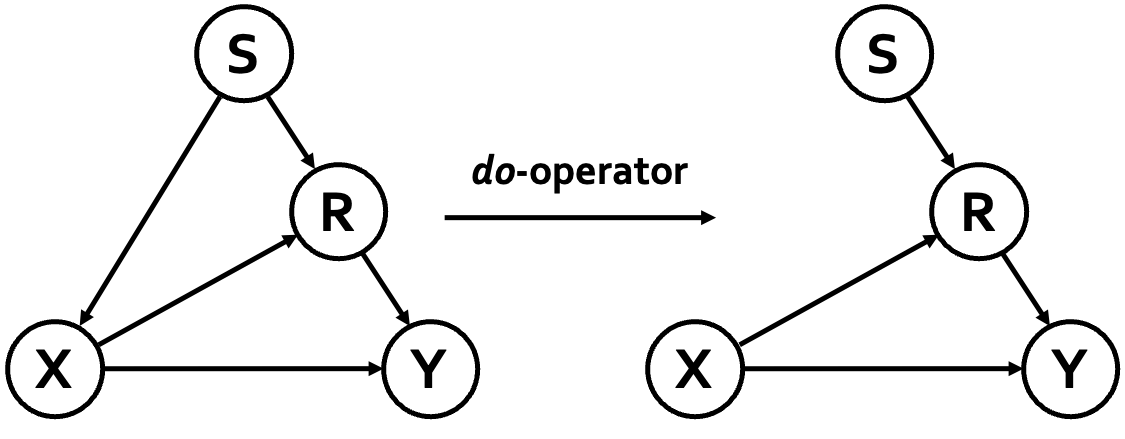}
    \caption{Illustration of our AU causal diagram.}
    \label{fig:causal_diagram}
\end{figure}

\begin{figure*}[!t]
    \centering
    \includegraphics[scale=0.54]{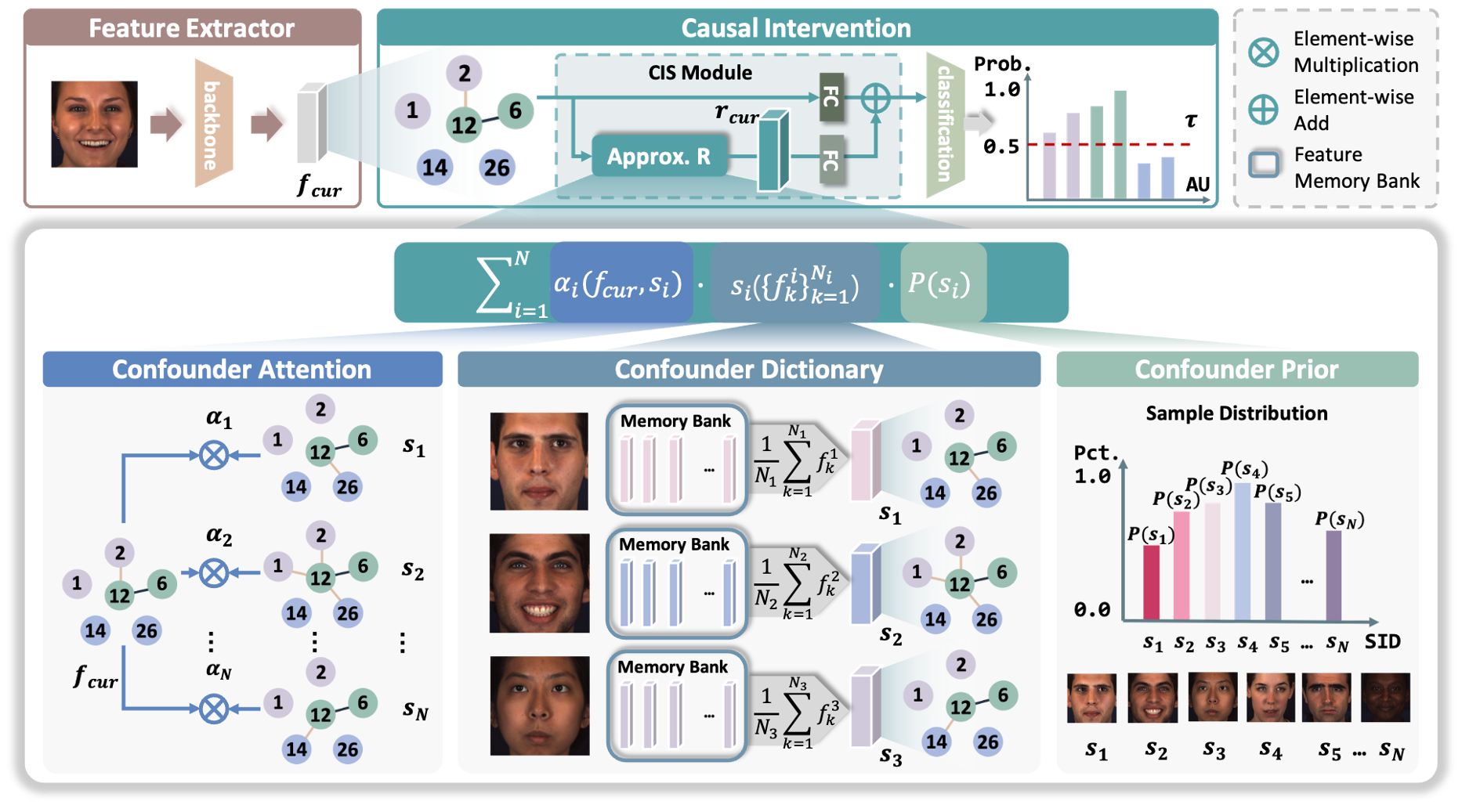}
    \caption{Overview. First, a facial image is fed into a backbone network for feature extraction. Instead of directly using the extracted feature $f_{\rm cur}$ for classification, we put $f_{\rm cur}$ into the proposed CIS module for causal intervention on \emph{Subject}, \ie~approximation of $P(Y|do(X))$. In CIS module, the output of \emph{Approx. R}---$r_{\rm cur}$ and $f_{\rm cur}$ are further fed into a linear layer separately and concatenated as the input of a classifier for AU prediction. The key component in CIS module is the approximation of $R$, which involves three parts for calculation, confounder attentions, a confounder dictionary, and confounder priors.}
    \label{fig:overview}
\end{figure*}

\section{Methodology}

\subsection{AU Causal Diagram}
To answer the whys and wherefores of subject variation problem, we use a structural causal model~\citep{pearl2000models}  to illustrate the causalities among variables in AU recognition models. As shown in Fig.~\ref{fig:causal_diagram}, there are four variables involved in our AU causal diagram, which are facial images $X$, subjects $S$, latent AU semantic relations $R$, and estimated AU occurrence probabilities $Y$, and the causalities among them are formulated via causal links, \ie~the direct edges in the causal diagram, each of which denotes the causality between two nodes, \ie~cause $\rightarrow$ effect. Causalities in our causal diagram are described in detail.

\subsubsection{$S\rightarrow X$} Subjects' customs of expressing emotions lead to subject-specific facial expressions in the facial images recorded while they are expressing emotions. In this way, $S$ determines what facial expressions appear on a subject's face, and for the same kind of facial expression, the facial appearance changes differ among subjects in a subtle but not negligible way. 

\subsubsection{$S\rightarrow R\leftarrow X$}
$R$ denotes latent AU semantic relations, which consist of both the universal AU semantic relations and the subject-specific AU semantic relations. The universal ones are determined by the facial anatomical basis of AUs, which is universal to all subjects. However, due to the custom differences in the way that subjects express emotions, not only the universal AU semantic relations but also the subject-specific AU semantic relations can be observed in the facial images recorded from one subject.
The latent AU semantic relations embedded in facial images are reflected by the causal link from $X$ to $R$. The learned AU semantic relations are embedded in a pre-trained AU recognition model, which contains subject-specific ones and can be reflected by the causal link from $S$ to $R$.

\subsubsection{$X\rightarrow Y\leftarrow R$}
Conventional AU recognition models aim to estimate AU occurrence probabilities $Y$ as precisely as possible. From our causal diagram, we can see that $Y$ is the effect of two causal paths, which are $X \rightarrow Y$ and $R \rightarrow Y$. The first causal path denotes that an AU recognition model estimates $Y$ based on the facial appearance features extracted from the input facial image, and the second one denotes that the learned latent AU semantic relations embedded in the model influence the estimated $Y$ by making use of priors from the training data. 
In other words, conventional AU recognition models which approximate $P(Y|X)$ learn a set of latent AU semantic relations $R$ from the training data, which can be regarded as a kind of priors influencing the estimation results $Y$ in an implicit way.
Although the effect brought by $R$ may introduce priors from the training data for better estimation when the facial appearance features are not sufficient enough to determine the states of certain AUs, it is confounded by subjects for training and thus may mistakenly associate or disassociate certain AUs.

\subsection{Causal Intervention via Back-door Adjustment}
To remove the adverse effect brought by confounder $S$ and obtain a model which estimates $Y$ only based on what's in $X$, \ie~facial appearance features of the input facial images, we propose to intervene $X$ by applying $do$-operator to variable $X$. 
The $do$-operator erases all the arrows that come into X, and in this way it prevents any information about X from flowing in the non-causal direction.
In this way, the causal link from $S$ to $X$ is cut-off, and we obtain an AU recognition model approximating $P(Y|do(X))$ instead of $P(Y|X)$.

The straightforward way to intervene $X$ is conducting a randomized controlled trial by collecting any facial image of any subject, and in this circumstance, $P(Y|X)$ equals $P(Y|do(X))$. Since such kind of intervention is impossible due to the infinity of the number of subjects and facial images, we apply the back-door adjustment formula as described in~\citep{pearl2016causal} to obtain $P(Y|do(X))$ in a simpler and feasible way.
To do this, we first estimate the effect at each stratum of the deconfounder in the intervention and then compute a weighted average of those strata, where each stratum is weighted according to its proportion. To be specific, the deconfounder is \emph{Subject} in our case, and we first estimate the causal effect for each subject in the training data and then estimate the average causal effect by computing a weighted average based on the proportion of each subject's facial images in the training data.

As described above, the effect of the intervention, $P(Y|do(X))$, can be formulated as follows:
\begin{equation}
    P(Y|do(X))=\sum_{s}P(Y|X,R=f(X,s))P(s), 
\label{eq:back-door}
\end{equation}
where $f(\cdot)$ is a function with $X$ and each $s$ as independent variables and $R$ as dependent variable. As $S$ is no longer correlated with $X$, the causal intervention makes $X$ have a fair opportunity to incorporate every subject $s$ into the estimation of $Y$, based on the proportion of each $s$ in the whole. After that, the causal link from $S$ to $X$ is cut off, which allows the causal effect $X \rightarrow Y$ free from the effect of $S$.

\subsection{CISNet Architecture}

\subsubsection{Overview}
As shown in Fig.~\ref{fig:overview}, our CISNet takes one facial image $X$ as input, and a backbone network is first applied to the facial image for feature extraction. Then the extracted feature $f_{\rm cur}\in \mathbb{R}^{d_{\rm in}}$ is passed through our plug-in causal intervention module, \ie~CIS module, for subject deconfounding. After that, the output of CIS module is fed into a classifier for AU recognition. By given a threshold $\tau$, the estimated AU occurrence probabilities as the classifier's output are processed as the final binary AU prediction results. 

\subsubsection{CIS Module}
We propose a plug-in causal intervention module, named CIS, for the approximation of the theoretical back-door adjustment formula.
As the calculation of Eq.~\ref{eq:back-door} requires the forward steps for each pair of $X$ and $s$, which cost a lot. Thanks to the Normalized Weighted Geometric Mean (NWGM) mentioned in~\citep{xu2015show}, which allows us to approximate the above expectation in feature-level.
In this way, our aim turns out to be computing Eq.~\ref{eq:NWGM}.
\begin{equation}
    P(Y|do(X))\overset{\rm{NWGM}}{\approx}P(Y|X,R=\sum_{s}f(X,s)P(s)).
\label{eq:NWGM}
\end{equation}

We apply a linear model to approximate the conditional probability, \ie~the probability of $Y$ under the conditions $X$ and $R$, as shown in Eq.~\ref{eq:linear_model}:
\begin{equation}
    P(Y|do(X)) = W_{\rm X}f_{\rm cur}+W_{\rm S}r_{\rm cur},
\label{eq:linear_model}
\end{equation}
where $W_{\rm X}\in \mathbb{R}^{d_{\rm out}\times d_{\rm in}}$ and $W_{\rm S}\in \mathbb{R}^{d_{\rm out}\times d_{\rm in}}$ are learnable weight parameters, $r_{\rm cur}\in \mathbb{R}^{d_{\rm in}}$ is an approximation of $R$.

The key component in CIS module is the calculation of $r_{\rm cur}$, which involves three parts, confounder attentions, a confounder dictionary, and confounder priors. 
For confounder $S$, since we cannot collect samples from all the subjects, we approximate $S$ as a fixed confounder dictionary $S=\left[s_1, s_2, \dots, s_N\right]$, where $N$ is the total number of subjects contained in the training data and each $s_i\in \mathbb{R}^{d_{\rm in}}$ is the subject prototype of \emph{Subject $i$}. To compute subject prototypes, we maintain one feature memory bank for each subject to store latent features, \ie~the output of the backbone network $\{f^i_{k}\in \mathbb{R}^{d_{\rm in}}\}^{N_i}_{k=1}$, for training samples of the certain subject, where $N_i$ is the number of samples for \emph{Subject $i$}. And in this way, $s_i$ is computed as $\frac{1}{N_i}\sum_{k=1}^{N_i}{f_k^i}$. 
Our confounder dictionary is updated as the end of each epoch, and confounder prior $P(s_i)$ can be computed as the ratio of the number \emph{Subject $i$}'s samples to the total number of samples for training. 
We approximate $R$ as a weighted aggregation of all subject prototypes, as shown in Eq.~\ref{eq:approx_r}:

\begin{equation}
    r_{\rm cur} = \sum_{i=1}^N{{\alpha_i}{s_i}{P\left(s_i\right)}},
\label{eq:approx_r}
\end{equation}

where $\alpha_i$ is a confounder attention for $s_i$ in the confounder dictionary with specific $f_{\rm cur}$, which is computed using Scaled Dot-Product Attention~\citep{vaswani2017transformer} as shown in Eq.~\ref{eq:attention}.

\begin{equation}
    \alpha_i = \operatorname{softmax}\left(\frac{{\left(W_{\rm Q}f_{\rm cur}\right)^T}\left(W_{\rm K}{s_i}\right)}{\sqrt{d_{\rm m}}}\right),
\label{eq:attention}
\end{equation}

where $W_{\rm Q}\in \mathbb{R}^{d_{\rm m}\times d_{\rm in}}$ and $W_{\rm K}\in \mathbb{R}^{d_{\rm m}\times d_{\rm in}}$ are learnable weight parameters.

\subsubsection{Loss Function}
To relief data imbalance problem in AU recognition, we apply an adaptive loss function for training:
\begin{equation}
    \mathcal{L}=-\sum_{i=1}^{C}\left[\left(1-\mu_{i}\right)p_{i}\log\hat{p}_{i}+\mu_i\left(1-p_{i}\right)\log\left(1-\hat{p}_{i}\right)\right],
\label{eq:loss}
\end{equation}
where $C$ is the number of AUs, $p_i$ is the ground-truth binary label for the $i^{\rm th}$ AU, and $\hat{p}_i$ denotes the estimated occurrence probability of the $i^{\rm th}$ AU. The occurrence frequency of the $i^{\rm th}$ AU in the training set, denoted by $\mu_{i}$, is maintained adaptively to weight the loss of each AU.

\section{Experiments}

\subsection{Datasets and Metrics}
In our experiments, we use two AU benchmark datasets, BP4D~\citep{bp4d} and DISFA~\citep{disfa}. BP4D involves 41 young adults, including 23 female and 18 male adults. Each subject is asked to finish 8 tasks, and 324 videos containing around 140,000 images are captured. Each frame is annotated with binary AU occurrence labels by two FACS coders independently. 
DISFA involves 26 adults, and to record their spontaneous facial behaviors, they are asked to watch specific videos. Each frame is annotated manually by a FACS coder with AU intensity labels within a scale of 0 to 5 for each frame. Frames with AU intensity labels greater than 1 are selected as positive samples. 

For each dataset, a subject-exclusive 3-fold cross-validation is conducted, following the experiment settings mentioned in~\citep{EAC-Net,SRERL_li2019semantic,song2021hybrid} for a fair comparison. 
We evaluate the proposed method using F1-score, defined as the harmonic mean between precision and recall and commonly used for multi-label classification.

\begin{table*}[!ht]
\centering
\begin{tabular}{l|cccccccccccc|c}
    \toprule
    Method & AU1   & AU2   & AU4   & AU6   & AU7   & AU10   & AU12  & AU14 & AU15 & AU17  & AU23  & AU24  & \textbf{Avg.}  \\
    \midrule
    DRML
    & 36.4 & 41.8 & 43.0 & 55.0 & 67.0 & 66.3 & 65.8 & 54.1 & 33.2 & 48.0 & 31.7 & 30.0 & 47.7 \\
    EAC-Net
    & 39.0 & 35.2 & 48.6 & 76.1 & 72.9 & 81.9 & 86.2 & 58.8 & 37.5 & 59.1 & 35.9 & 35.8 & 55.6 \\
    ROI-Net
    & 36.2 & 31.6 & 43.4 & \underline{77.1} & 73.7 & \underline{85.0} & 87.0 & 62.6 & 45.7 & 58.0 & 38.3 & 37.4 & 56.3 \\
    DSIN
    & 51.7 & 40.4 & 56.0 & 76.1 & 73.5 & 79.9 & 85.4 & 62.7 & 37.3 & 62.9 & 38.8 & 41.6 & 58.9 \\
    JAA-Net
    & 47.2 & 44.0 & 54.9 & \textbf{77.5} & 74.6 & 84.0 & 86.9 & 61.9 & 43.6 & 60.3 & 42.7 & 41.9 & 60.0 \\
    LP-Net
    & 43.4 & 38.0 & 54.2 & \underline{77.1} & 76.7 & 83.8 & \underline{87.2} & 63.3 & 45.3 & 60.5 & 48.1 & 54.2 & 61.0 \\
    SRERL
    & 46.9 & 45.3 & 55.6 & \underline{77.1} & \textbf{78.4} & 83.5 & \textbf{87.6} & 63.9 & \textbf{52.2} & \underline{63.9} & 47.1 & 53.3 & 62.9 \\
    UGN-B
    & \underline{54.2} & \underline{46.4} & \underline{56.8} & 76.2 & 76.7 & 82.4 & 86.1 & 64.7 & 51.2 & 63.1 & \underline{48.5} & 53.6 & 63.3 \\
    HMP-PS
    & 53.1 & 46.1 & 56.0 & 76.5 & \underline{76.9} & 82.1 & 86.4 & \underline{64.8} & \underline{51.5} & 63.0 & \textbf{49.9} & \underline{54.5} & \underline{63.4} \\
    \midrule
    CISNet
    & \textbf{54.8} & \textbf{48.3} & \textbf{57.2} & 76.2 & 76.5 & \textbf{85.2} & \underline{87.2} & \textbf{66.2} & 50.9 & \textbf{65.0} & 47.7 & \textbf{56.5} & \textbf{64.3} \\
    \bottomrule
\end{tabular}
\caption{F1-score (\%) for 12 AUs reported by the proposed method and the state-of-the-art methods on BP4D dataset. The best and second results are indicated using bold and underline, respectively.}
\label{tab:f1_bp4d}
\end{table*}

\begin{table*}[!ht]
\centering
\begin{tabular}{l|cccccccc|c}
    \toprule
    Method & AU1 & AU2 & AU4 & AU6 & AU9 & AU12 & AU25 & AU26 & \textbf{Avg.} \\
    \midrule
    DRML
    & 17.3 & 17.7 & 37.4 & 29.0 & 10.7 & 37.7 & 38.5 & 20.1 & 26.1 \\
    EAC-Net
    & 41.5 & 26.4 & 66.4 & 50.7 & 8.5 & \textbf{89.3} & 88.9 & 15.6 & 48.5 \\
    DSIN
    & 42.4 & 39.0 & 68.4 & 28.6 & 46.8 & 70.8 & 90.4 & 42.2 & 53.6 \\
    JAA-Net
    & 43.7 & 46.2 & 56.0 & 41.4 & 44.7 & 69.6 & 88.3 & 58.4 & 56.0 \\
    LP-Net
    & 29.9 & 24.7 & \underline{72.7} & 46.8 & \underline{49.6} & 72.9 & \underline{93.8} & \underline{65.0} & 56.9 \\
    SRERL
    & \underline{45.7} & 47.8 & 59.6 & 47.1 & 45.6 & 73.5 & 84.3 & 43.6 & 55.9 \\
    UGN-B
    & 43.3 & \underline{48.1} & 63.4 & 49.5 & 48.2 & 72.9 & 90.8 & 59.0 & 59.4 \\
    HMP-PS 
    & 38.0 & 45.9 & 65.2 & \underline{50.9} & \textbf{50.8} & 76.0 & 93.3 & \textbf{67.6} & \underline{61.0} \\
    \midrule
    CISNet 
    & \textbf{48.8} & \textbf{50.4} & \textbf{78.9} & \textbf{51.9} & 47.1 & \underline{80.1} & \textbf{95.4} & \underline{65.0} & \textbf{64.7} \\
    \bottomrule
\end{tabular}
\caption{F1-score (\%) for 8 AUs reported by the proposed method and the state-of-the-art methods on DISFA dataset. The best and second results are indicated using bold and underline, respectively.}
\label{tab:f1_disfa}
\end{table*}

\subsection{Implementation Details}
For each input image, Dlib~\citep{Dlib} is used to detect facial landmarks. According to the computed coordinates of eye centers, we align the image, crop the facial region, and resize the cropped face to $256\times 256$. 
We employ ResNet34~\citep{resnet} without the final linear layer ($512\!\rightarrow\!64$, $64\!\rightarrow\!C$) as our backbone network, and two linear layers act as the classifier. $d_{\rm in}$ and $d_{\rm out}$ are set to 512 and $d_{\rm m}$ is set to 256. Before training, we set $N$ feature memory banks to save features for each subject in the training data, and $N$ equals 27, 27, 28 for each fold respectively. In the first epoch, we stop the back-propagation step to initialize our confounder dictionary. Threshold $\tau=0.5$ is used for the binarization of the predicted probabilities $\hat{p}\in \mathbb{R}^{C}$.

CISNet is implemented on PyTorch~\citep{pytorch} platform. We use Stochastic Gradient Descent (SGD) with momentum of 0.9 and weight decay of 0.0005 as the optimizer. Learning rate is set to 0.001 and batch size is set to 4. The number of training epochs is set to 15, and early stopping strategy is employed for training. All models are trained on one NVIDIA Tesla V100 16GB GPU.

\subsection{Comparison with State-of-the-art Methods}
We compare the performance of CISNet with the previous state-of-the-art methods including DRML~\citep{DRML_2016_CVPR}, EAC-Net~\citep{EAC-Net}, ROI-Net~\citep{ROINet_2017_CVPR}, DSIN~\citep{DSIN_2018_Corneanu}, JAA-Net~\citep{JAA-Net}, LP-Net~\citep{LP_2019_CVPR}, SRERL~\citep{SRERL_li2019semantic}, UGN-B~\citep{song2021uncertain} and HMP-PS~\citep{song2021hybrid}.

As shown in Table~\ref{tab:f1_bp4d}, CISNet outperforms all the compared methods in terms of average F1-score on BP4D. Specifically, CISNet achieves an average F1-score of $64.3\%$, which outperforms HMP-PS with the second-highest F1-score by $0.9\%$.
Table~\ref{tab:f1_disfa} shows the experimental results on DISFA. CISNet outperforms HMP-PS by a large margin with the highest average F1-score of $64.7\%$.
It is worth mentioning that the proposed CIS module can be inserted into almost all frame-based AU recognition models suffering from subject variation problem for causal intervention.

\begin{table}[!b]
\centering
\setlength{\tabcolsep}{0.8mm}{
\begin{tabular}{lccccccc}
\toprule
 & \multicolumn{3}{c}{BP4D} & \multicolumn{3}{c}{DISFA} \\
\midrule
         & w/o CIS & w/ CIS  & $\Delta$      & w/o CIS & w/ CIS & $\Delta$ \\
ResNet18 & 59.9 & \textbf{63.8} & 3.9$\uparrow$ & 57.3 & \textbf{63.5} & 6.2$\uparrow$\\
ResNet34 & 60.6 & \textbf{64.3} & 3.7$\uparrow$ & 57.4 & \textbf{64.7} & 7.3$\uparrow$\\
ResNet50 & 61.3 & \textbf{63.6} & 2.3$\uparrow$ & 56.2 & \textbf{65.0} & 8.8$\uparrow$\\
\bottomrule
\end{tabular}}
\caption{Effectiveness of CIS on different backbones.}
\label{tab:different_backbones}
\end{table}

\subsection{Ablation Study}
Our ablation studies aim to answer: \textbf{Q1.}~How does CIS module perform in models with different backbone networks? 
\textbf{Q2.} Is the number of training subjects the larger the better?

\subsubsection{A1. Effectiveness of CIS on Different Backbones}
To better elucidate the effectiveness of CIS, we evaluate the improvement brought by CIS module inserted in models with different backbone networks including ResNet18, ResNet34 and ResNet50~\citep{resnet} in Table~\ref{tab:different_backbones}. By inserting CIS into models with different backbone networks, we can observe significant and consistent improvements compared with the corresponding models w/o CIS, which owes to the reduction of prediction bias via deconfounding of \emph{Subject}.

\begin{figure}
    \centering
    \includegraphics[scale=0.45]{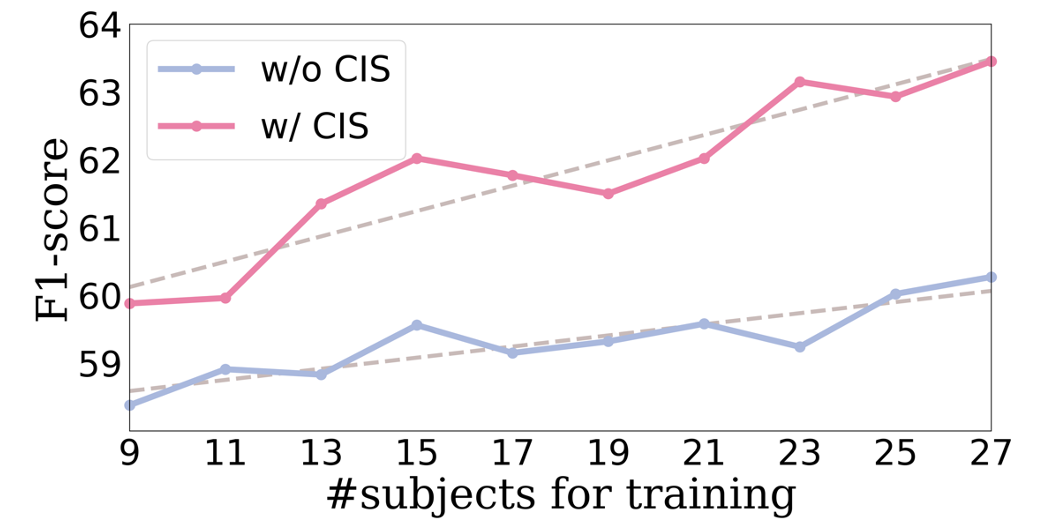}
    \caption{Impact of the number of training subject.}
    \label{fig:subject_sufficiency}
\end{figure}

\subsubsection{A2. Impact of the Number of Training Subjects}
Considering that back-door adjustment is based on the premise of sufficient data, we conduct experiments to study the impact of the number of training subjects by training a baseline model (w/o CIS) and CISNet (w/ CIS), which only differ in terms of CIS module. Fig.~\ref{fig:subject_sufficiency} shows that with the number of training subjects increasing, the performance of two models on the same test data increases accordingly, but the increasing rate of CISNet is slightly higher than the baseline model, which illustrates the importance of a sufficient number of training subjects and the potential of performance improvement brought by CIS module with more training subjects. 

\begin{figure}[!ht]
    \centering
    \includegraphics[scale=0.37]{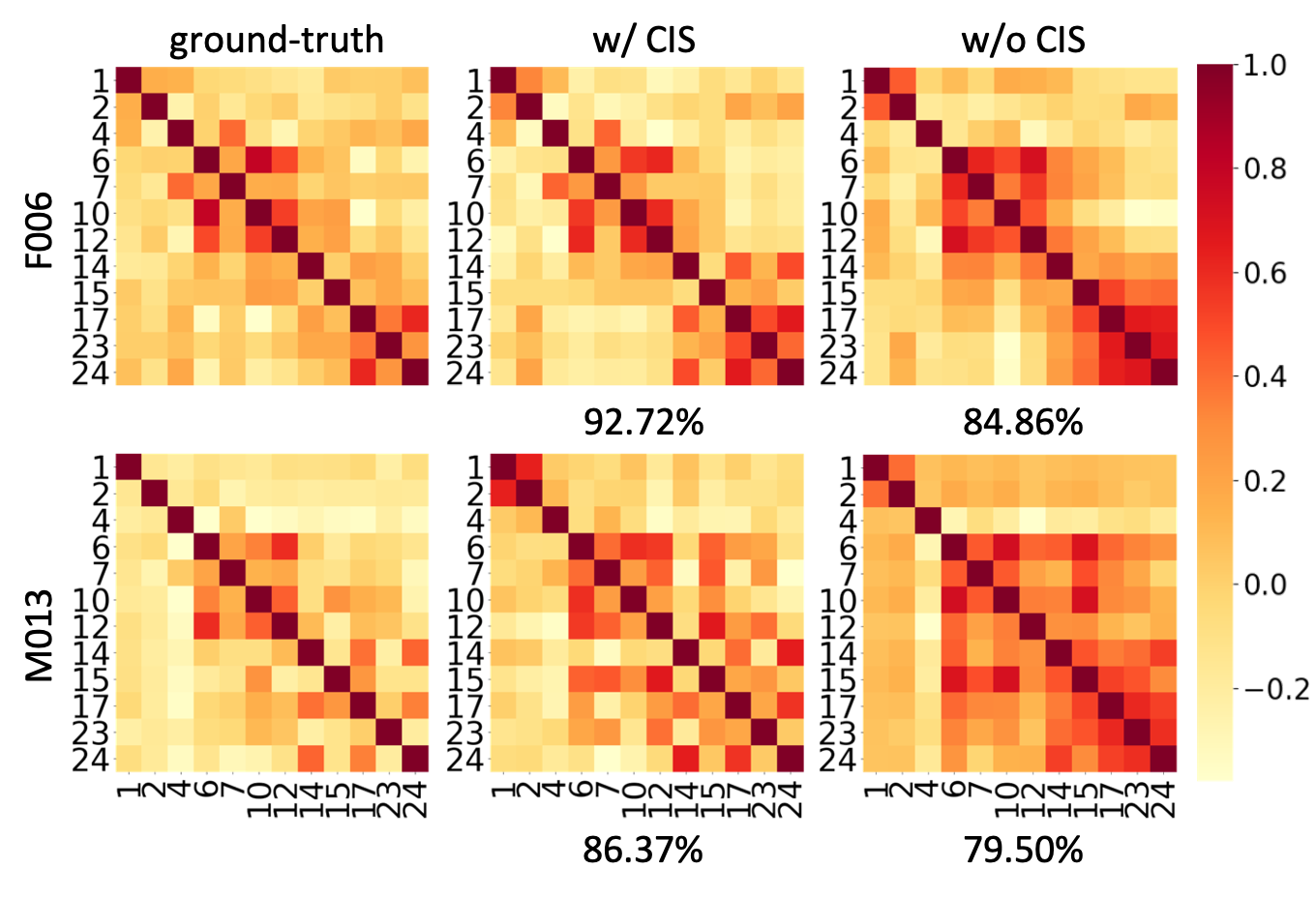}
    \caption{PCC among AUs for different subjects. From left to right, PCC matrices are computed based on the ground-truth AU labels, predicted ones using CISNet (w/ CIS), and predicted ones using the baseline model (w/o CIS), respectively. Numbers under PCC heatmaps are cosine similarities between themselves and the corresponding ground-truth.}
    \label{fig:pcc_heatmap}
\end{figure}

\subsection{Qualitative Results}
To better explain the mechanism of CIS module, we provide qualitative results in this section to answer:
\textbf{Q3.}~Does CIS module truly assist the model to estimate $Y$ only based on what's in $X$? \textbf{Q4.}~Does the AU representations extracted by our CISNet invariant to subjects? \textbf{Q5.}~What's the differences between models approximating $P(Y|X)$ or $P(Y|do(X))$?

\subsubsection{A3. Visualization of PCC Heatmaps}
To illustrate that CIS module acts as a causal intervention on \emph{Subject} and makes the model estimate $Y$ only based on $X$ without unnecessary or even harmful prior from the training data, we visualize the Pearson Correlation Coefficient (PCC) matrices computed based on ground-truth or predicted AU labels for each subject.
Fig.~\ref{fig:pcc_heatmap} shows that the PCC heatmaps computed for the baseline model and CISNet differ from each other, and the PCC heatmaps for CISNet are more similar to the PCC heatmaps computed based on the ground-truth AU labels according to their cosine similarities, which demonstrates that by using CIS to deconfound \emph{Subject}, the model is endowed with the capability to focus on what's in $X$ from a new subject without the confounding of latent subject-specific AU semantic relations implied in the training data.

\subsubsection{A4. Visualization of Representations}
\begin{figure}[!ht]
    \centering
    \includegraphics[scale=0.31]{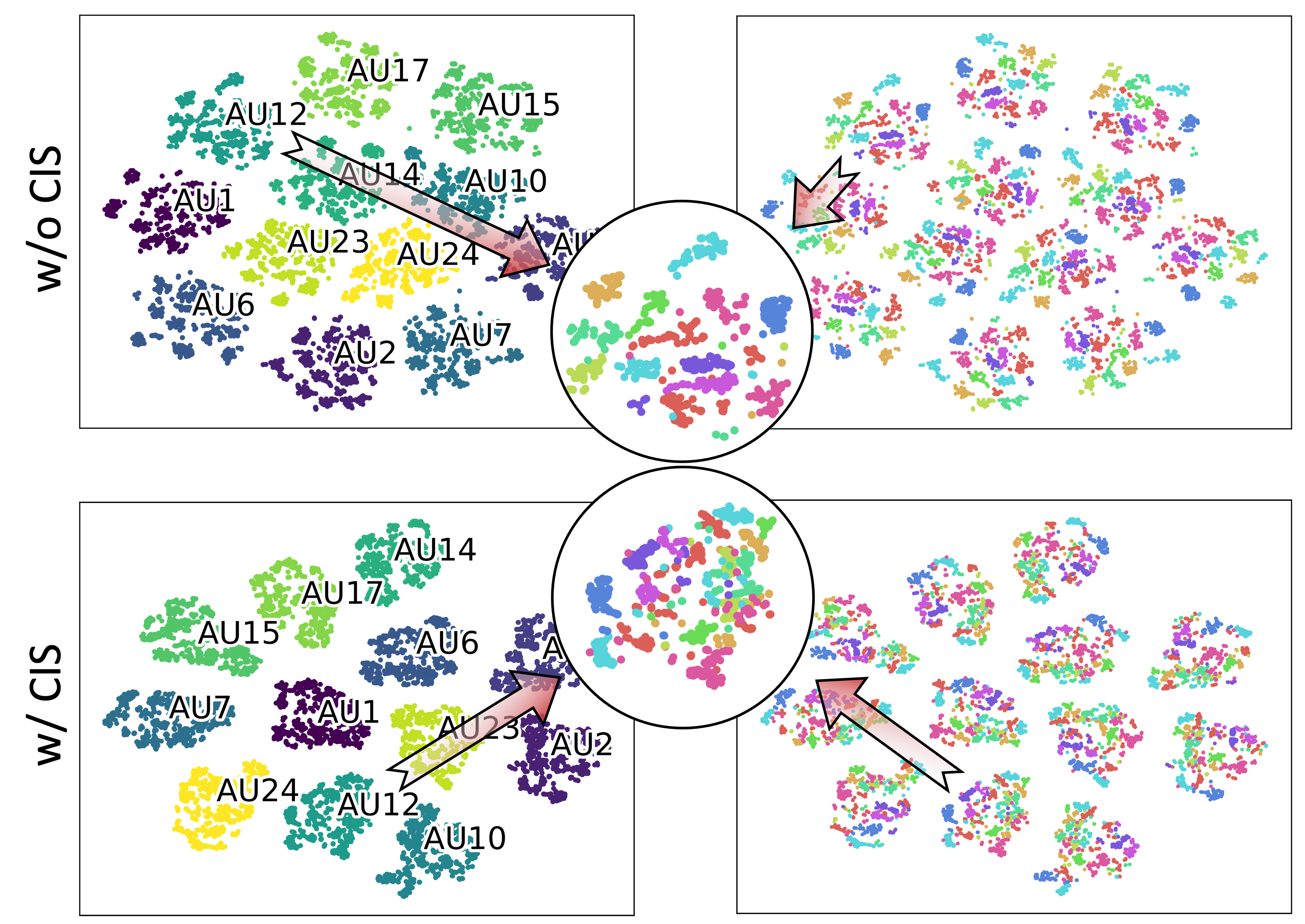}
    \caption{t-SNE visualization for AU-specific vanilla backbone w/ or w/o CIS module on BP4D dataset.}
    \label{fig:tsne}
\end{figure}

To show that the representations learned by CISNet are more subject-invariant, we insert $C$ spatial-attention layers~\citep{zhao2019pyramid} between the backbone network and the classifier to obtain AU-specific features, where $C$ is the number of AUs, and use t-SNE~\citep{tsne} for visualization.
As shown in Fig.~\ref{fig:tsne}, in the visualization result of the baseline model, features with the same subject labels form small clusters in a cluster corresponding to a specific AU. As for CISNet, features with the same subject labels are more dispersed in a cluster corresponding to a specific AU, which illustrates that the representations learned by CISNet are more invariant to subjects.

\begin{figure}[!ht]
    \centering
    \includegraphics[scale=0.41]{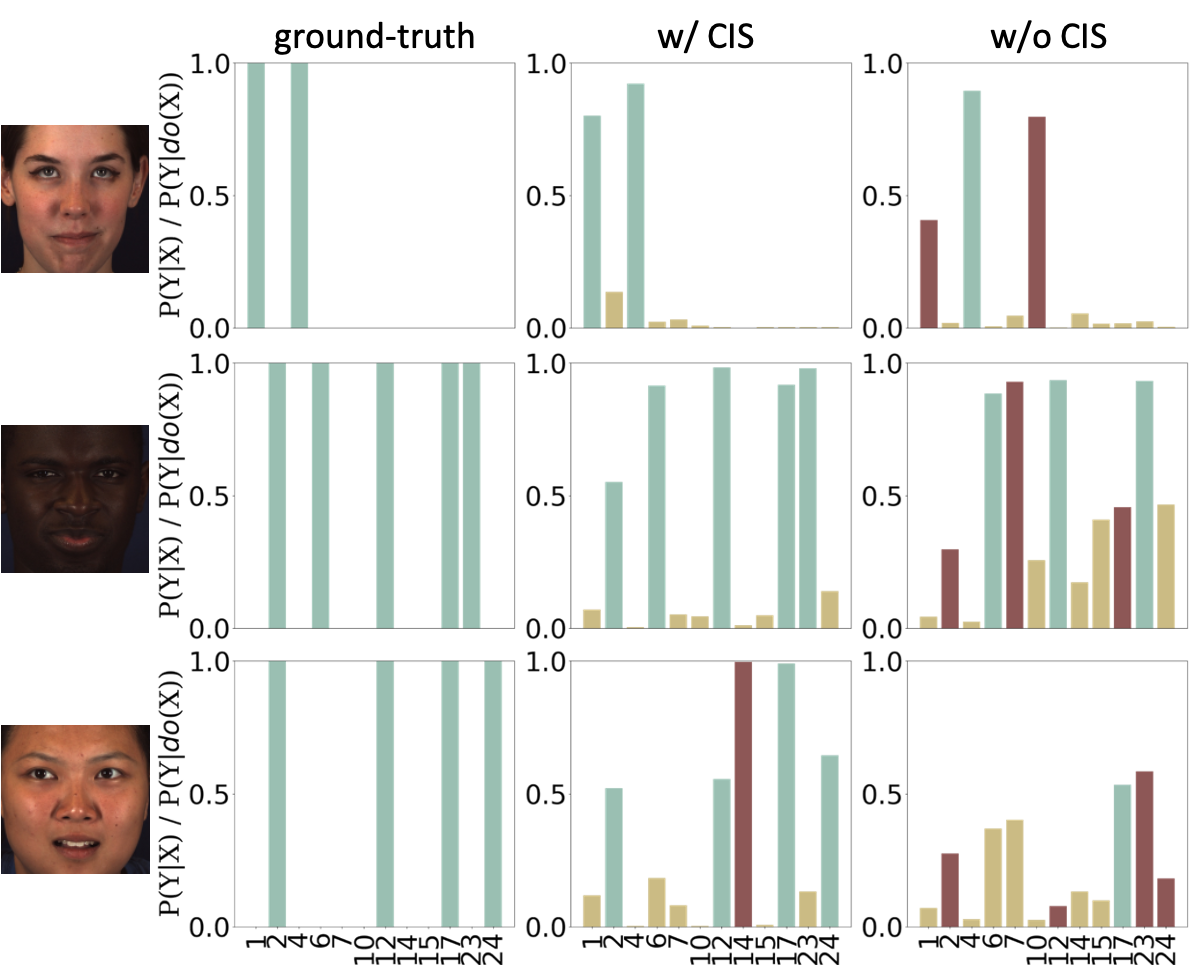}
    \caption{Differences between $P(Y|X)$ and $P(Y|do(X))$.}
    \label{fig:case_study}
\end{figure}

\subsubsection{A5. Case Study}
We visualize the estimated AU occurrence probabilities of several samples to show the differences between $P(Y|X)$ and $P(Y|do(X))$. From Fig.~\ref{fig:case_study} we can see that the probabilities estimated by CISNet are more close to the ground-truth, and the probabilities estimated by the baseline model reflect some AU semantic relations such as the co-occurrence of AU6 and AU7 which are not suitable for subjects in the samples. Such kind of prior from the training data leads to poor prediction results during inference.

\section{Conclusion}
This paper focuses on explaining the why and wherefores of subject variation problem in AU recognition with the help of causal inference theory and providing a solution for subject-invariant facial action unit recognition by deconfounding variable $S$ in the causal diagram via causal intervention. Unlike previous works that made attempt to deal with this problem through subject-specific learning or domain adaptation, we proposed a plug-in causal intervention module named CIS to remove the adverse effect brought by confounder \emph{Subject} in a straightforward way, which could be inserted into almost all frame-based AU recognition model and boost them to a new state-of-the-art. Extensive experiments prove the effectiveness of our CIS module, and vanilla backbones with CIS module inserted achieve state-of-the-art results.

\section*{Acknowledgments}
This work is in part supported by the PKU-NTU Joint Research Institute (JRI) sponsored by a donation from the Ng Teng Fong Charitable Foundation.

\bibliography{references}

\end{document}